# A Novel Topology for End-to-end Temporal Classification and Segmentation with Recurrent Neural Network

Taiyang Zhao
sonosole@163.com

**Abstract**

Connectionist temporal classification (CTC) has matured as an alignment free to sequence transduction and shows competitive for end-to-end speech recognition. In the CTC topology, the blank symbol occupies more than half of the state trellis, which results the spike phenomenon of the non-blank symbols. For classification task, the spikes work quite well, but as to the segmentation task it does not provide boundaries information. In this paper, a novel topology is introduced to combine the temporal classification and segmentation ability in one framework.

## 1 Introduction

A continuous speech recognition system trained with a neural network requires training targets for every segment in the sequential input for supervised training, which means the alignment between inputs and outputs is required to provide the targets. CTC achieves this by introducing a blank symbol to form an ingenious topology which maps the input sequences to the labelled sequences, thus the alignment of the labels with the inputs is not important. In the CTC topology, the blank symbol occupies more than half of the state trellis, which results the localized peaks of the non-blank symbols in a sequence. For classification task, the peaks prediction works quite well, but as to the segmentation task it does not provide precise location information. Actually, in some tasks like keyword spotting, sound event detection and protein secondary structure prediction, segmentation is as important as classification. In [1], bottleneck features from the model trained with CTC are clustered into either background cluster or foreground cluster using unsupervised clustering method, but the inaccuracy of the clustering limits its usage and it is a two-stage method. The attention mechanism deterministically performs alignment between acoustic features and recognized symbols [2], however the basic temporal attention mechanism is too flexible that it allows non-sequential alignments, which is not proper in speech recognition task where the acoustic features and the corresponding outputs are aligned in a monotonic way.

In this paper, we combine the temporal classification and segmentation (TCS) ability in one framework with the use of a new topology. Based on this topology, one can use the same training criteria as CTC for end-to-end TCS task. In addition, all the strategies to make CTC training stable can be applied to TCS training such as gradient clipping [3], and SortaGrad method [4].

## 2 Topologies

In the topology of CTC as shown in figure 1 (a), the blank symbol always has one more state than all other symbols in a sequence. When summing all blank posteriors at a given time step in the forward-backward procedure, the target for blank state will be much higher than all other states. As a result, the network trends to go through paths with many blanks. As the training continuous, the network would output blanks only, but predicting only blanks at every time step would increase the cost since any correct path must contain all non-blank symbols. As a compromise, the network has to find one location to predict each symbol. Finally, the outputs of the network have a pretty steep posterior probability distribution between blank and non-blank symbols. Predicting very localized peaks for symbols is interesting in decoding phase. Since the blank is not very informative, most blanks could be skipped which accelerates the decoding speed. CTC fails when it comes to segmentation issue, because it cannot give the boundaries of the characters. [5] tried to avoid the peaks of predictions by enforcing each valid path contains several consecutive repetitions of the same character before being allowed to transit to a blank. But when they repeated each character $n$ times in training phase, the length of each character during decoding was almost always $n$.

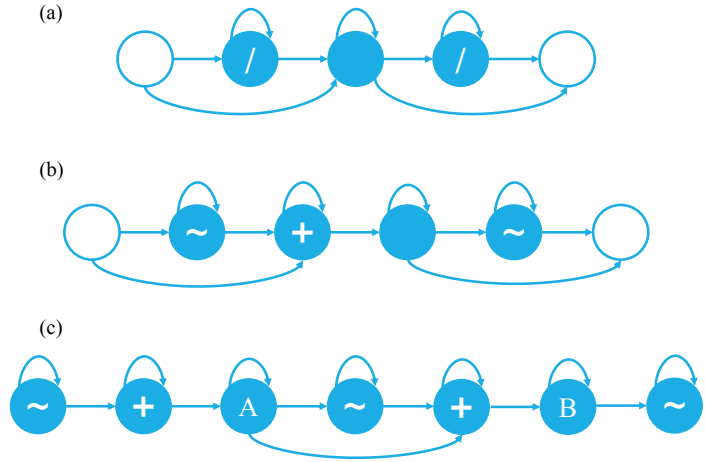

Figure 1: (a) CTC topology, (b) TCS topology and (c) TCS topology for sequence label 'AB'. The symbol '/' stands for blank label, '~' for background label and '+' for foreground label

From these observations, we attempted to design a new topology to reduce the steepness of the posterior probability distribution between different states while remaining the peak phenomenon. Our topology is shown in figure 1(b). Every valid character shall only be emitted after a foreground state, then before and after every pair of foreground label and character label there is an optional background label. If the length of the original label sequence is $U$, the length of the modified label sequence is therefore $3U+1$. Figure 1 illustrates all possible paths that maps input sequence in the target sequence 'CAT'. The final result we need is the sequence which does not have foreground and background labels. We can simply do this by removing first the repeated labels and then the foreground and background labels from the paths. For example, F(~~++C++AA~+T~) = F(~++C++A++T) = CAT. Once the target posteriors are computed by the forward-backward algorithm [6], gradients of the cross entropy loss between the softmax outputs and the targets are back propagated through the network.

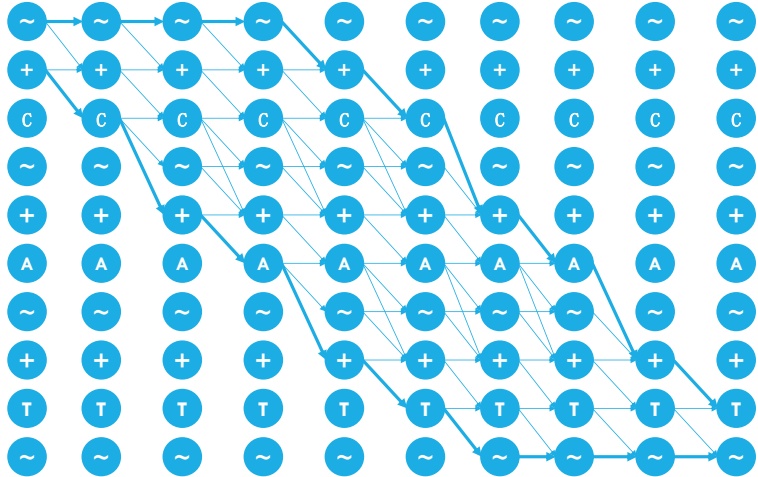

Figure 2: Trellis of the labelling 'CAT'.

## 3  Experiments and Result

In a 0-to-9 digital number recognition task, we use 32-dimentional log mel-filterbank energy features computed every 8ms on 16ms windows. To apply sufficient context and reduce the computation, 8 contiguous frames are stacked into a super-frame every 2 frames. The network consists of 3 hidden recurrent layers, where each layer has 128 hidden units. The last layer is a softmax layer with 12 units (one unit for the background label, one for the foreground label, and the rest for syllable labels). In figure 3, TCS is applied a speech signal labelled '127934'. As we expected, the background state is just aligned with the non-speech segment, the foreground state is aligned with the speech segment, and after each speech segment there follows a speech state. This clear correspondence, on the one hand, reduces the learning difficulty of the RNNs, on the other hand, it makes it possible to explicitly skip the non-speech segments when decoding online, which is more convincing than the way CTC skips most blank states. What is more, the proposed method is humanly understandable, the network must wait until sufficient information is acquired before giving an identification, just like a human understands what a linguistic pronunciation means as long as it was reaching the end. The disadvantage is that TCS consumes more calculations than CTC during the training phase.

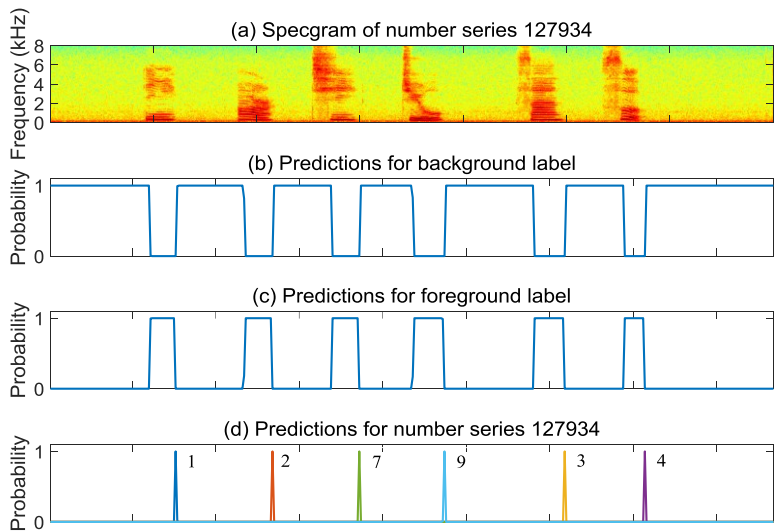

Figure 3: TCS applied to a speech signal labelled '127934'.

# 4 Conclusion

This paper proposed a novel topology to combine the temporal classification and segmentation ability within one framework. Although we haven't done many experiments to prove that TCS is better than the existing end-to-end system, but the digital number recognition task has demonstrated the promise of TCS based temporal modeling methodology.